# THEORY-BASED INDUCTIVE LEARNING: AN INTEGRATION OF SYMBOLIC AND QUANTITATIVE METHODS

Spencer Star[*]

Department of Computer Science, Carnegie-Mellon University
Pittsburgh, PA 15213   Arpanet: Spencer.Star@H.CS.CMU.EDU

***ABSTRACT:*** *The objective of this paper is to propose a method that will generate a causal explanation of observed events in an uncertain world and then make decisions based on that explanation. Feedback can cause the explanation and decisions to be modified. I call the method Theory-Based Inductive Learning (T-BIL). T-BIL integrates deductive learning, based on a technique called Explanation-Based Generalization (EBG) from the field of machine learning, with inductive learning methods from Bayesian decision theory. T-BIL takes as inputs (1) a decision problem involving a sequence of related decisions over time, (2) a training example of a solution to the decision problem in one period, and (3) the domain theory relevant to the decision problem. T-BIL uses these inputs to construct a probabilistic explanation of why the training example is an instance of a solution to one stage of the sequential decision problem. This explanation is then generalized to cover a more general class of instances and is used as the basis for making the next-stage decisions. As the outcomes of each decision are observed, the explanation is revised, which in turn affects the subsequent decisions. A detailed example is presented that uses T-BIL to solve a very general stochastic adaptive control problem for an autonomous mobile robot.*

## 0. Introduction

Suppose that a decision is to be made when the state of the world is uncertain but further information about it can be obtained by measurement or experimentation. It is given that the decision must be logically consistent with the decision maker's own preferences for consequences, as expressed by numerical utilities, and with the weights he attaches to the possible states of the world, as expressed by probabilities. This is the classical problem of decision analysis (Raiffa and Schlaifer 1961; Savage 1954).

If an analyst is asked *why* he believes a decision to be correct or why the observed outcome occurred, he can offer an explanation that relates theories, facts, and the context. Although explanation facilities have long been considered as central to the construction of expert systems in AI, it is only recently that attempts have been made to integrate the explanation facility with the basic theory-driven inductive approach of decision analysis (Langlotz et al. 1986).

In machine learning, the term explanation has been given a precise meaning in the context of *explanation-based generalization* (EBG), a technique used to generate an explanation of a concept after observing a single training example (Mitchell et al. 1986). An explanation in EBG is a proof, based on prior knowledge, that an observed event is a deductive implication of a set of propositions. The explanation is generalized to cover not only the observed instance but a more general class of instances.

An explanation in EBG serves quite a different purpose than an explanation produced by an explanation facility in an expert system. In an expert system, an explanation is used to show the *user* the reasoning process behind a conclusion. The explanation has been entered by a programmer into the expert system and is not used by the system's inference procedures. In EBG, the explanation has not been written into the system by a programmer but is generated by the system's own inference procedures. If a subsequent activity requires using knowledge contained in the explanation, the system's inference procedures will have access to and can manipulate the explanation. It is as if the program can explain to *itself* what it has learned.

Learning programs in AI have not, as a rule, been able to deal gracefully with noisy data, and EBG is no exception. The problem that these programs have in dealing with stochastic influences probably is related to the origins of the approach in a tradition that emphasizes the use of deductive logic. EBG can trace its roots to the goal regression techniques developed for the STRIPS robot problem-solving system by Fikes et al.(1972). From those beginnings to the present day, learning programs from the AI tradition have tended to limit their domain to non-stochastic learning and have put the emphasis on symbolic computation.

Learning that takes place in a stochastic environment has been studied most often in fields that are outside of the mainstream of AI work. The historical roots are different from those of AI, the journals are different, and the practitioners are often in other academic departments. Stochastic learning is central to much of the work done in pattern recognition, stochastic adaptive control theory, dynamic programming, Bayesian decision theory, and statistical inference. The emphasis in these areas is on quantitative methods as opposed to symbolic computation.

In this paper symbolic methods will be integrated with quantitative methods to develop a procedure that can (1) generate a causal explanation of observed events, (2) make decisions based on that explanation, and (3) use feedback to revise the explanation and decisions. This requires solving two problems which I call *the explanation problem* and *the decision problem*. The explanation problem requires the generation and revision of a probabilistic causal explanation relating the state of the world to uncertain outcomes that are the consequences of actions initiated by the decision maker. The decision problem requires using the explanation to make a logical choice among alternative courses of action in the decision maker's uncertain world.

## 1. A Brief Review of Explanation-Based Generalization (EBG)

As presented by Mitchell, Keller, and Kedar-Cabelli

---

[*]Currently on leave from Université Laval, Départment d'informatique, Québec, Canada. I would like to thank Paul Black, Oren Etzioni, Max Henrion, Steve Minton, Tom Mitchell, and Prassad Tadepalli for their many helpful comments.

(1986), EBG is a technique for learning by generalizing explanations. The system is shown a single training example of an event. It first constructs a specific explanation for the event, and then it generalizes this explanation so that it covers a class of similar events.

The EBG method requires the following information:

(1) *Goal Concept*: A concept definition describing the concept to be learned. An example used by Mitchell et al. (1986) of a goal concept is Safe-to-Stack(x,y). The goal was to determine for a pair of objects <x,y> if it is safe to stack x on top of y. The objects were a box and a table, and the initial definition used the predicate, weight(x). It was assumed that there was no direct way to measure the weight of the box, but that its density and volume could be measured. The final generalized explanation therefore used the terms volume(x) and density(x) along with the relationship between volume, density, and weight.

(2) *Training Example*: A specific example of the concept in terms of observable features. An instance of a training example for the goal Safe-to-Stack(x,y) would be that a box with given density and volume can be safely stacked on a table with given weight and sturdyness.

(3) *Domain Theory*: A set of rules and facts sufficient to prove that the training example is an example of the goal concept.

(4) *Operationality Criterion*: A requirement that the final generalized concept definition be described in terms of the predicates used to describe the training example or in terms of a selected set of easily evaluated predicates from the domain theory. It is assumed that the initial definition includes at least one unobservable term, thus failing to satisfy the Operationality Criterion.

Given this information, an EBG system will construct a specific explanation of why the training example satisfies the goal concept. The specific explanation is a logical proof that demonstrates that the event is entailed by prior information in the form of Horn clauses in the knowledge base. The proof is generalized by propagating constraints on the variable bindings among the various proof rules, but dropping any constraints introduced by the specific example. The EBG technique for generalizing uses a form of goal regression that has been shown to be equivalent to an augmented version of resolution theorem proving for Horn Clause Logic (Fikes et al. 1972; Waldinger 1977; Nilsson 1980; Mitchell et al. 1986; Mooney and Bennet 1986; Kedar-Cabelli and McCarty 1987).

When EBG has constructed the valid generalization from the training example, the system is said to have learned the goal concept. The kind of learning done by using EBG can be characterized as being *truth preserving* and *nonampliative*. By truth preserving I mean that there is no possibility that a conclusion about an event will be false. By nonampliative I mean that nothing is learned that was not previously, at least implicitly, known to the system. In EBG, since all events that could possibly be observed can be explained by the available domain theory and all explanations use valid forms of deductive logic, the explanation is only an explicit formulation of knowledge that was already implicitly available. Any learning that has taken place is similar to learning that occurs when one completes a mathematical proof of a theorem.

In contrast to the learning done by EBG, inductive learning is not truth preserving but it is ampliative. The generalizations and predictions that are based on inductive inferences can be wrong, which is the price we pay in order to learn something new about the world.

Since all the knowledge to generate the final goal concept is available in the knowledge base, why is a training example used? Why not just let the system generate the proof tree of the final goal concept from the initial definition and the domain knowledge? The standard reply has been that the use of a training example and generalization procedure makes the search more efficient. There is, however, another reason that I would like to propose as being significant: The training example provides a contextual constraint so that the generalized proof structure is also a meaningful and relevant explanation of the goal concept.

If EBG is to be used when there is a great deal of domain knowledge, it is likely that there will be more than one valid proof structure that entails the goal concept. For example, the goal concept Safe-to-Stack(x,y) could include clauses stating that x is not made of glass and has a flat bottom surface while y has a flat top surface and is level. If our objective is to determine if x can be stacked on y without crushing y, and if x and y will always be cardboard boxes of similar size but different weights in the problem context, then although the facts and theory concerning flat surfaces and glass containers could be part of a valid proof structure, they are irrelevant to the desired goal concept. EBG will omit them from the explanations it generates.

The problem of finding the appropriate explanation clearly involves more than merely finding a valid relation between theory and fact: It depends on a three-term relation between theory, fact, and context (van Fraassen 1980). An explanation can be considered to be the answer to a why-question about the occurrence of a certain event. Why is it safe to stack object x on object y? A satisfactory answer depends on the context in which the question was asked. EBG solves the problem of determining context by requiring a training example to be presented to the system and by using a generalization technique that follows the specific proof actually used in the explanation of the training example (DeJong and Mooney 1986). It ignores the other possible ways to complete a proof. The use of a single training example to determine context is one of the major benefits that EBG brings to theory-based inductive inference.

We have seen that EBG can generalize and determine context on the basis of a single training example. Its efficiency at generalization has been bought at the cost of requiring substantial domain knowledge about the world. There appears, however, to be another associated cost: The capacity to do inductive inference has seemingly been eliminated. This has led researchers to the opinion that an important next step in the machine-learning research program is to try to integrate empirically-based inductive methods with EBG (Mitchell et al. 1986; Lebowitz 1986). The remainder of this paper will propose how this might be done.

## 2. Combining Probabilistic Reasoning, Utilities, and EBG

EBG has several important limitations that must be relaxed before it can be applied to generate and modify explanations used in T-BIL. First, EBG builds ex-planations using purely deductive, non-probabilistic logic. We need to allow probabilistic and statistical reasoning to enter into its explanations. Second, although EBG constructs explanations and learns concepts, the concept is learned in a vacuum. I am interested in concepts that are learned because they will have a certain utility for a future decision. The goal concept must therefore be linked to a decision with observable and measurable consequences. Third, in EBG no method has been proposed for revising an explanation on the basis of new information. A method for incremental updating of the existing explanation is needed. Fourth, EBG cannot represent disjuncts in its explanations. A technique for including disjunctive propositions will be introduced.

Probabilistic reasoning will be integrated with EBG by associating probablities with every proposition in the explanation. Propositions will be referred to as hypotheses if the probability of their being true is less than one. Definitions will have probability one that they are true. Probabilities are to be interpreted as quantitative measures of belief in the truth of a hypothesis. Prior probabilities are conditional on the context and background conditions which are only made explicit to a partial degree in an explanation. Where possible, probabilities will be given a objective interpretation; otherwise they are subjective probabilities. Incremental updating will be done by applying Bayes' theorem.

The combination of probabilistic reasoning, utilities, and EBG to do theory-based inductive learning should be referred to by a unique name in order to maintain the distinctions between it and the other methods for doing inductive inference, learning, and decision analysis. I will call the method *Theory-Based Inductive Learning*, and will refer to it simply as T-BIL.

T-BIL requires information that is substantially different than EBG in order to solve problems and generate explanations. Instead of providing an initial definition of a goal concept, T-BIL requires an initial statement of a *decision problem*. In its most general form, the decision problem requires sequential decisions to be made when the state of the world, and thus the outcome of any decision, is uncertain. Moreover, it is assumed that additional information can be obtained at a cost, either by performing an experiment or by some other means. Both EBG and T-BIL require a *domain t eory*. There is no essential difference in the way this term is used in EBG and T-BIL.

T-BIL uses a *training example*, as does EBG, as the basis for its generalization procedure. In EBG we can consider the initial goal concept to provide the boundary conditions for the most general version of the concept and the training example to provide the boundary conditions for the most specific version of the concept. Goal regression constructs the bridge between those two boundaries, finding the right mix of specialization and generalization. In T-BIL, goal regression will accomplish the same task, specializing from the initial statement of the decision problem and generalizing from the training example to find the right level of explanation that links the two. This is the major contribution that symbolic methods from machine learning make to T-BIL.

The contributions that quantitative methods make to T-BIL are associated with the use of Bayesian decision theory. The decision maker using T-BIL observes the outcome of his initial decisions and uses this information to update beliefs and revise decisions. Instead of referring to these additional observations as training examples, I will call them *observation reports*. The distinction is maintained, therefore, between the training example, which is used with goal regression to generate an explanation, and observation reports, which are used with Bayesian updating to modify the explanation.

In EBG the operationality criterion is a restriction that forces the generalization to use terms that were used in the training example. In T-BIL, I would like to use a concept of operationalism similar to that first proposed by the Nobel physicist P.W. Bridgman (1927). The idea as adopted by many scientists is that the terms in a theory be reducible to operations, usually involving observation and measurement, that can be publicly repeated. There must also be a high degree of agreement among the observers as to the outcome of the operation. In essence, theoretical explanations have meaning only through their connection with the observable. Since this is different from the operationality criterion of EBG, I have called the requirement an *operational explanation*.

Before discussing how to combine decision analysis with EBG, let us look more closely at the decision problem. The general decision problem that interests us is dynamic and stochastic: It involves decisions over time that transform one uncertain state into another. The decision maker chooses acts so as to maximize the sum of the expected utilities over time. It is assumed that the time involved is not long enough to require using discounted present values. The chosen sequence of decisions is called a policy or a plan. A plan that permits the decision maker to maximize the sum of the expected utilities is called an optimal plan. This description of the decision problem casts it as a problem in stochastic adaptive control theory (Bellman 1961; Dreyfus and Law 1977).

More formally, the decision problem of finding the optimal plan can be described as a dynamic maximization problem

$$\max \sum_{i=0}^{N} EU_i(X_i, D_i) \quad \text{s.t.} \quad X_{i+1} = T_i(X_i, D_i)$$

where the variables and relevant features of the decision problem are defined as follows:

(i) $D_i$: Decision set at time i. The finite set of potential acts by the decision maker. A decision maker can select a specific act {d} from the set of possible acts $D_i = \{d1, d2, ..., dn\}_i$.

(ii) $X_i$: State of the world at time i. A state refers to the finite set of feasible outcomes of the world. Each outcome has associated with it a joint probability measure.

(iii) $U_i$: Utility evaluation in the state $X_i$. The decision maker assigns utility $U(d, x)$ for each decision and each feasible outcome. Probability times utility is expected utility, $EU(X, D)$.

(iv) $P(X_i/d_i)$: Probability assessment on $X_i$. For every decision $d_i$ the decision maker directly or indirectly assigns a joint probability measure $P(X_i/d_i)$ to each possible outcome in the state at time i.

(v) *T*: Transformation rules. A set of rules that map each element of $X_i$ into an element of $X_{i+1}$.

The terms and the information required for T-BIL are presented in more detail as follows:

*(1) Decision Problem:* A problem that requires the logical analysis of choice among courses of action from the decision set, $D_i$, when the outcome of any act depends on the current state of the world, $X_i$, which is known with uncertainty, and it is possible at a cost to obtain additional information about the current state. The decision maker's preferences for consequences are expressed by his utility evaluation $U_i$, and the weights he attaches to possible outcomes are expressed by his probability assessment $P(X_i/d_i)$.

*(2) Training Example:* A specific example of a correct solution to the decision problem. The training example should contain information that is sufficient, given the assumed background conditions, to solve at least one stage of a sequential decision problem.

*(3) Domain Theory:* A set of rules and facts sufficient to prove that the training example is a solution to one stage of the decision problem. The domain theory either includes the necessary prior probabilities and assumptions about the conditional independence of various hypotheses or can obtain them from another information source.

*(4) Observation Reports:* Reports to the decision maker on the outcome of measurements, experiments, queries, or decisions that provide additional information relevant to the decision problem. Observation reports can contain the same data as a training example, but a training example is an instance used in the generalization process while an observation report is an instance used to revise probabilities associated with the already-generalized explanation.

*(5) Operational Explanation:* A requirement that the terms used in the generalized explanation be reducible to terms that are observable and measurable in the problem contexts. Theoretical explanations have meaning only through their connection with the observable. Thus decision procedures, the training example, and observation reports must ultimately be based on observable and measurable terms.

Given this information in (1) - (4) and the constraint in (5), the system will generate a specific explanation to justify why the training example solves the initial decision problem. This explanation is structured as an influence diagram that is formed by using the inference rules and the conditional independence relations in the domain theory. The next step is to generalize this explanation, using goal regression. The generalized explanation consequently covers a class of instances while specialized explanation only covers the training example. The decision maker will make a decision, observe the outcome, and revise his beliefs, thus moving from one state to the next.

### 3. A Brief Review of Influence Diagrams

In EBG, the explanation structure is a proof tree formed using first-order logic. In order to represent causal explanations that involve probabilistic measures of uncertainty, the proof tree will be replaced by a graph called an influence diagram (Howard and Matheson 1981; Shachter 1986; Pearl 1986). Readers having a background in decision theory are undoubtedly familiar with influence diagrams; but those whose background is AI may find this brief reveiew of influence diagrams useful.

An influence diagram is an acyclical directed graph having two types of nodes: *chance nodes* and *decision nodes*. A chance node corresponds to a random variable and is drawn as a circle. A decision node is drawn as a rectangle.

Any well-formed influence diagram can be transformed into a decision tree. Decision trees rapidly become unwieldy, so that in complex problems they do not provide as clear, concise, or intuitive an explanation as influence diagrams. Moreover, influence diagrams provide an explicit representation of probabilistic dependencies that are only implicit in decision trees.

*Conditioning influences* are represented by arrows entering chance nodes. These arrows show the variables on which the probability distribution of the chance node variable will be conditioned. They indicate conditional dependence. In the diagram below, variable C depends on B and B depends on A, but C is conditionally independent of A. This means that P(C/B)=P(C/B,A). In other words, any effect that A has on C occurs through its effect on B. If there were another arc directly connecting A and C, then the possibility would exist that C is not conditionally independent of A. It is preferable to omit the arc between two nodes that are conditionally independent because the graph then explicitly represents the independencies.

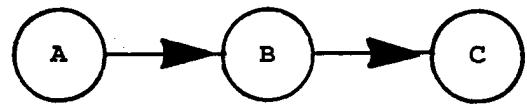

An important consequence of Bayes' theorem is that arcs that connect chance nodes can be reversed without making any incorrect or additional assertions about the possible independence of variables provided that all probability assignments are based on the same set of information (Howard and Matheson 1981). Bayes' theorem permits one to calculate P(A ) or P(B/A) indifferently given the same information.

Since arcs between chance nodes only indicate the possibility of conditional dependence, it is possible to add arcs between any pair of chance nodes without changing the meaning of the diagram. One cannot, however, remove an arc unless conditional independence is proposed.

*Informational influences* are represented by arrows leading into a decision node. They represent a basic cause-effect ordering. In the following diagram the arrow from the chance node D to the decision node E is an informational influence. This means that the information resulting from knowledge of the random variable D is available to E at the time of the decision. The arrow between D and E cannot be reversed without changing the state of information at the time of the decision. The arrow from decision node E to chance node F is a conditioning influence on F. It means that what occurs in F is conditionally dependent on the results in E. But the information in F is not known before the decision at E is taken.

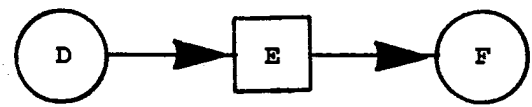

Influence diagrams are based on the supposition that a single person (or machine) is the decision maker. Another related assumption is the "no forgetting" rule. All information that corresponds to events that occur before a decision is made is available to the decision maker. He cannot "forget" that something happened.

Influence diagrams must be acyclical. It is sometimes the case that a directed cycle will be formed if a decision leads to successor chance nodes which have arrows leading into a chance node that is a predecessor to the decision. In that case it is usually possible using Bayes' theorem to reverse an arc between two chance nodes, thus eliminating the cycle.

## 4. A Robot's Choice-Of-Path Problem

### 4.1 Problem Statement

Dynamic programming is an optimization procedure that is applied to problems requiring a sequence of interrelated decisions. When the outcomes of the decisions are stochastic and there is feedback, the theory for solving these problems is frequently referred to as *stochastic adaptive control theory* or *Bayesian decision theory*. Very many of the problems associated with the techniques and theory in this area can be represented as a form of the shortest-path problem (Dreyfus and Law, 1977). The following *robot's choice-of-path problem* is simply a specific instance of the very general stochastic shortest-path problem. It has been structured to be a further development of the Safe-to-Stack(x,y) example used by Mitchell et al. (1986).

Suppose that we have available an autonomous mobile robot that can do a variety of tasks. We would like the robot to follow a path from point A to point B using the shortest route, subject to certain constraints. Let us further suppose that the route traverses 200 intersections. At each intersection the robot can observe that the shortest path is blocked by a box and a table. The robot must decide if it can safely stack the box on the table. Its domain knowledge includes facts such as half of the boxes have density of .3 and half have density of .4, the table is believed to be Fragile with P(Fragile)=.80 or Sturdy with P(Sturdy)=.20, weight is equal to volume times density, a densimeter can measure a box's density with a random error, and specific utilities are attached to each outcome. In terms of the T-BIL method presented in previous sections, we can describe the robot's problem as follows:

*(1) Decision Problem:* Choose the PathDecision in each state that will maximize the decision maker's total utility of going from an initial point A to a terminal point B. It requires choosing either the long or short path at each intersection. The utility associated with each outcome is measured in arbitrary units called utils. If the short path is taken, the box is stacked on the table, and the table does not break, the outcome is worth +100 utils. If the table breaks, the outcome is worth -100 utils. If the long path is taken, the certain outcome is worth +10 utils.

```
Max[Utility(PathDecision(1)+...
       +PathDecision(200))]
Utils((ShortPath&Resists +100)
      (ShortPath&Breaks -100)
      (LongPath +10))
```

*(2) Training Example:* The ShortPath is chosen. Box-0 is stacked on Table-0, and the table Resists. Table-0 weighs 4 units. Box-0 has known volume and density.

```
IsA(Box-0 Box State(0))
Density(Box-0 .4 State(0))
Volume(Box-0 10 State(0))
IsA(Table-0 Table State(0))
Weight(Table-0 4 State(0))
Outcome(Table-0 Resists State(0))
PathChosen(Short-Path State(0))
```

*(3) Domain Theory:* Default values for box densities show that half the boxes have density of .3 and half have density of .4. The densimeter has a measurement error of 0 or plus or minus .1 with equal probability of occurrence. It is assumed that the class Table is either Fragile or Sturdy. The initial belief is .8 that all tables are Fragile and .2 that they are all Sturdy. If a table is Fragile, it will always support a box that weighs less than the table; it will resist without breaking a box that weighs the same as the table only 40% of the time. If a table is Sturdy, it will resist a box that weighs the same as the table 60% of the time. All tables to be encountered weigh 4 units. All boxes have volume of 10 units. The formula relating weight, volume, and density is known. Measurement of density can be take with a densimeter.

```
BoxDensities(P(.3)=.5 P(.4)=.5)
DensimeterError(P(.1)=.33
                P(0)=.33
                P(-.1)=.33)
Table(P(Fragile)=.8 P(Sturdy)=.2)
Fragile(P(Resists-Lighter-Wt)=1
        P(Resists-Equal-Wt)=.6)
Sturdy(P(Resists-Lighter-Wt)=1
       P(Resists-Equal-Wt)=.4)
Volume(Box P(10)=1)
Weight(Table P(4)=1)
Formula(Weight=Volume*Density)
Measurement(Density Densimeter)
```

*(4) Observation Reports:* Information on (1) whether the table Breaks or Resists when a box is stacked on it, and (2) the PathChosen.

*(5) Operational Explanation:* The causal explanation must be reducible to a description in terms of features that are observable and measurable in the problem context. For this problem, the following key terms are observable and measurable: Short-Path, Long-Path, Breaks, Resists, Density, Volume. The terms Sturdy and Fragile are not directly observable, but are defined in terms of frequencies of observable events. The Weight of boxes is not directly observable in the problem context, but is defined in terms of Density and Volume, which are observable.

Given this information, the system will generate an influence diagram, which is the explanation structure that shows why the training example is a solution to the decision problem. Figure 1 shows an influence diagram that corresponds to the specific explanation for the training example. Figure 2 presents the influence diagram that corresponds to the generalized explanation that exists in state 1, the situation when the robot arrives at the first intersection and must make a decision. In Figures 1 and 2, I have used a diamond shape to indicate the final outcome where utility is assigned.

241

It corresponds to a specialized chance node. Figure 3 presents the decision tree for state 1.

The influence diagram in Figure 2 shows the explanation with the context determined by the training example. The explanation can be used to answer why-questions involving the two decisions and the observed events:

- Why did you measure density?
- Why did you take the short path?
- Why did the table resist?

The explanation provides the directly relevant antecedents to the decisions and observed events. The context provided by the training example determines which of many possible explanations is chosen and where the explanation stops: what is relevant and, most importantly, what is irrelevant.

The decision problem is subject to combinatorial explosion. In our example, the robot will have to solve the path-choice problem 200 times. Thus at each terminal node in the decision tree there should be another decision tree. Since there are 15 non-zero probability terminal nodes, this would require $15^{200}$ terminal nodes. The approach used here follows the basic method of dynamic programming by breaking the large problem up into a multistage problem that can be solved one stage at a time. This reduces the number of paths to be explored from $15^{200}$ to a more manageable $15*200$.

In the following sections, the relationship between each figure and the underlying decision processes will be explained in more detail.

*4.2 The Specific Explanation*

Consider the specific explanation of the training example as represented in Figure 1. The nodes are numbered and will be referred to as N1, N2, and so on. Begin the analysis with the decision represented by N8. The decision to take the short path has the consequences that the table resisted the box's weight without breaking, thus leading to an outcome valued at 100 utils. The observation that the table resisted was furnished by the training example, while the utility value was inferred from the statement of the decision problem, which is supplied along with the training example. Starting at N7, the method examines the preconditions that must be fulfilled for the table to resist. The table can be either sturdy or fragile, but N5 is a terminal node so no further regression can be done. At N6, the box can weigh less than or the same as the table, but there are two nodes that determine the value of this node. Go first to N4, a terminal node, and find the weight of the table. At N3 the weight of the box is unknown, but it can be determined by the density and volume, which are given in the training example, along with the formula for weight, which is part of the domain knowledge. Although it would be possible to represent explicitly the weight formula, it has been omitted for conciseness. Once the values for nodes 1 and 2 are found, they are propagated down and the diagram corresponds to Figure 1.

*4.3 The Generalized Explanation*

Figure 2 is the influence diagram of the generalized explanation that is built from the specific explanation in Figure 1. The two major differences between the two figures are (1) the generalization from the instances of Box-0 and Table-0 to the classes Box and Table, and (2) the addition of three new nodes (N10, N11, N12) reflecting knowledge from the domain theory about measuring density. If the generalized explanation structure in Figure 2 is well formed, enough information exists to transform it algorithmically into a decision tree that can be used to solve the decision problem.

For Figure 2, let us consider first only the nodes that have been generalized, nodes 1 through 9. At N1, it is stated that all boxes have densities of either .3 or .4, and that these are uniformly distributed among the population of boxes. From N2, boxes have a default volume of 10, giving at N3 an equal distribution of weights being either 3 or 4. Since from N4 tables have a default weight of 4, N6 has the conditional probabilities

$P(BoxWeight=TableWeight/N3, N4) = .50$

$P(BoxWeight<TableWeight/N3, N4) = .50$.

N5 has two unconditional subjective probabilities about the belief that a table is Fragile or Sturdy, and two objective probabilities that are used in the definitions of Fragile and Sturdy. The subjective prior beliefs about whether the table is Fragile or Sturdy are

*P(Fragile) = .8*

*P(Sturdy) = .2.*

Note that we are representing the disjunct that tables are either Sturdy or Fragile. The objective probabilities associated with the definitions are

*P(Resists/Fragile, SameWeight) = .40*

*P(Breaks/Fragile, SameWeight) = .60*

*P(Resists/Sturdy, SameWeight) = .60*

*P(Breaks/Sturdy, SameWeight) = .40*

The probabilities from N5, and N6, and the decision in N8 lead to nine conditional probabilities in N7. Eight of these are conditional on a decision to take the short path plus the three other pairs of probabilities from N5 and N6, while one is conditional on a possible decision at N8 to take the long path.

Before a decision is considered to measure density, the nine probabilities that are calculated at N7 correspond to the nine terminal nodes of the c5 branch of the decision tree in Figure 3. The probabilities are multiplied by the appropriate utilities at N9, which are propagated up the decision tree to give the expected utility of 44 at the second-level decision c5 in Figure 3.

A decision must be made about whether or not to measure density before the path decision is made. If a box has a measured density of .5, the robot will be able to infer that the box weighs 4 units. If the measured density is .2, it will infer that the box weighs 3 units. Other measured densities provide no additional information over the basic information provided at N1.

The inclusion of the density-measurement decision will require more probabilities to be calculated for state 1. The robot will have to consider the probability of each density value conditional on the decision to measure at N10, the error distribution at N11, and the distribution of densities among boxes at N1. These four conditional probabilities are prior information that will become known with certainty at the time of the decision to choose a path. This is indicated by the informational arrow going from N12 into N8. This in turn requires four new sets of conditional probabilities to be calculated at N7, with outcomes at N9, conditional on each possible outcome of the measurement act. The appropriate probabilities can be found by tracing the upper branches of the decision tree from b1 to the terminal nodes. In state 1 there are 39 different paths through the decision tree, of which 24 terminate at some point with a zero probability of being taken.

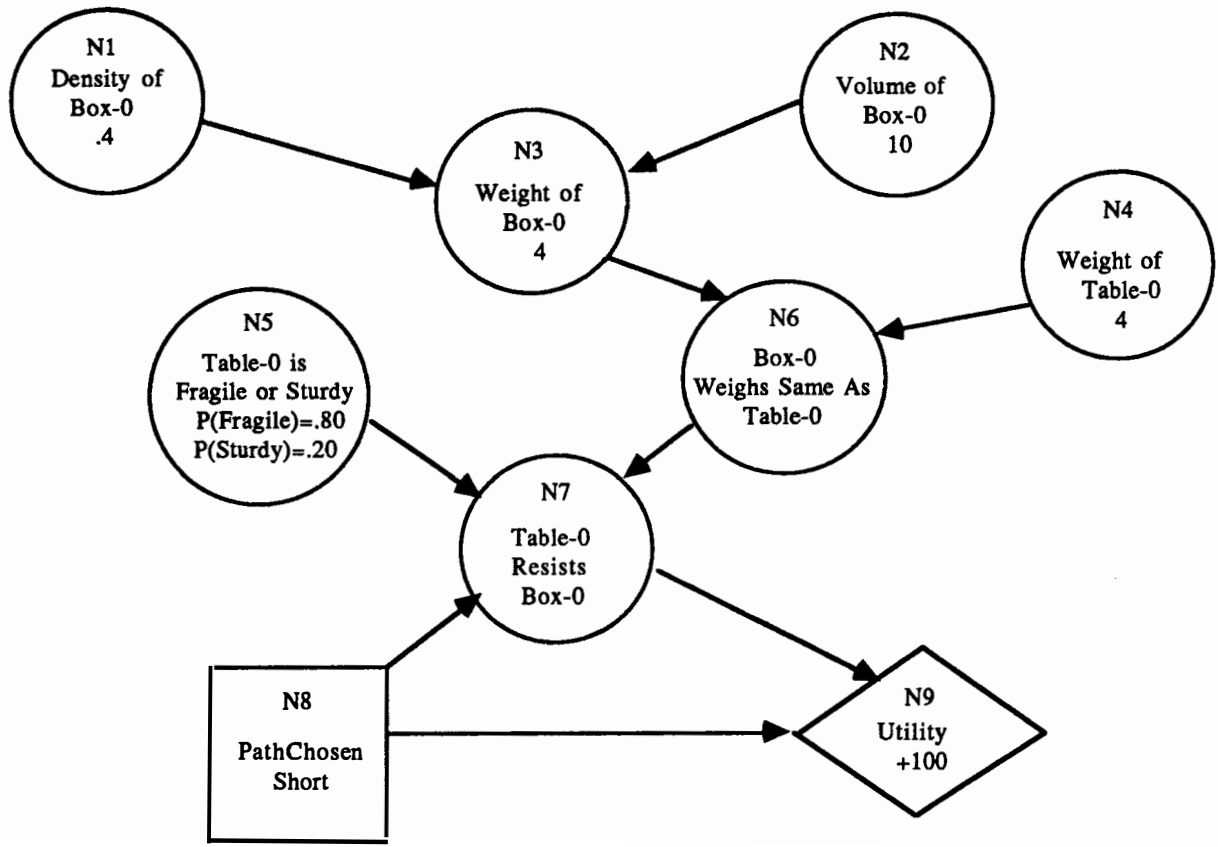

Figure 1. Specific Explanation of Training Example

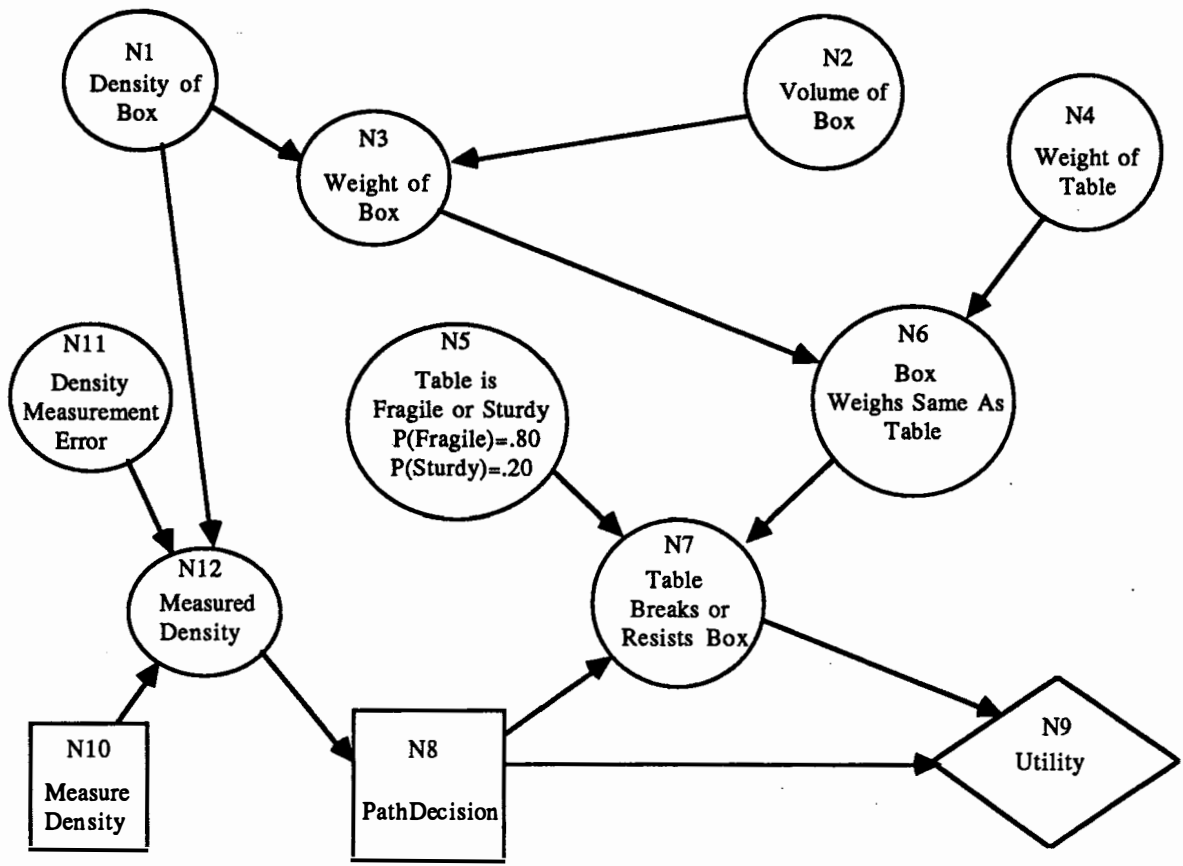

Figure 2. Generalized Explanation for Choice-Of-Path Problem

The value of perfect information can be easily calculated from the decision tree. Given perfect information about box densities, utility will be maximized if (a) the short path is taken whenever a box weighs three units and (b) the long path is taken whenever a box weighs four units. The values of this strategy correspond to values at nodes c1 and c4, each with associated probabilities of .5. The expected utility of this plan is 55 utils. The expected utility of making the path decision without density measurement is only 44 utils (c5). If density measurement is performed, the optimal strategy in state 1 will be to take the short path in all cases except when measured density is .5. This will give an expected utility of 47.7 utils. Thus perfect information adds 11 utils over the uninformed choice, while density measurement adds 3.7 utils. In our example, measurement costs nothing, but if its cost were equal or greater than 3.7 utils, it should not be done.

It is also possible to measure the value of perfect information about the supportive capacity of tables. First, suppose that the robot correctly knows that the tables are Sturdy. Then the optimal strategy will be always to take the short path. The realized utility will be +20 for heavy boxes and +100 for light boxes, yielding an average utility for perfect information of 60. Next, suppose that the robot knows correctly that the tables are Fragile. In this case, the current strategy in Fig. 3 is optimal, the utilities at d3 and d5 will be 40, and the realized utility for perfect information will be 45. The expected utility from perfect information will be the weighted sum of those two utilities, with weights depending on the robot's current beliefs in the strength of tables:

$$48 = .2(60) + .8(45)$$

Perfect information about the strength of tables is thus worth only 0.33 utils, the difference between what he believes perfect information is worth (48.00 utils) and the expected value of his current strategy (47.67 utils).

A summary of what has been accomplished up to this point may be useful. First, we have solved one part of the explanation problem: A generalized explanation in the form of an influence diagram has been constructed from a problem statement and a training example. The explanation can be used to answer why-questions involving each decision and the observed consequences of the decision: Why was it decided to ...?, and Why was it observed that ....? The explanation uses concepts that sometimes are not directly observable, such as the weight of a box or the expected utilities of an action, but the explanation is grounded in observable features, such as density, volume, table breaks, measured density, etc. Second, we have used the generalized explanation to construct a decision tree corresponding to state 1. Although a decision tree can always be derived from a well-formed influence diagram, it is not strictly necessary in order to solve the decision problem. It is possible to go from the influence diagram directly to the solution, with the decision tree being only an implicit intermediate step. The decision tree was used to explain the details of the solution to the decision problem in state 1. The value of perfect information and of a decision to gather more information was inferred from the available information, and a decision was made to measure density and to choose a path conditional on this information. It remains to consider how the explanation and decisions are modified by inductive learning as the robot progresses from one state to another.

*4.4 Inductive Learning Between States*

The passage between states occurs as a result of a standard series of transformation rules which can be summarized as *decide, observe,* and *update*. The previous section discussed how to decide. The problem now is to observe and update so that inductive learning takes place. In order to illustrate how this is done, let us suppose that the box to be encountered at intersection 1, Box-1, has a measured density of .4, that the short path is taken and Box-1 is stacked on Table-1, and that the table resists the weight of the box. How would these observations change the robot's beliefs?

The inductive learning is manifested by a change in beliefs at N5 about whether the table is Fragile or Sturdy. The odds-ratio form of Bayes' rule will prove to be convenient for discussing inductive learning. Form the ratio of the two beliefs using the standard form of Bayes' equation to get the odds ratios for the case of the table resisting:

$$O_{post} = L \cdot O_{prior}$$

where the posterior odds are

$$O_{post} = \frac{P(Fragile/Stacked, Resists)}{P(Sturdy/Stacked, Resists)},$$

the prior odds are

$$O_{prior} = \frac{P(Fragile)}{P(Sturdy)},$$

and the likelihood ratio is

$$L = \frac{P(Resists/Fragile, Stacked)}{P(Resists/Sturdy, Stacked)}.$$

To calculate the value of the likelihood ratio when a box is stacked and the table resists, let us consider 100 boxes and assume that they are perfectly divisible. If we start with 100 boxes, 50 heavy and 50 light, we eliminate the 16.67% of the boxes that are heavy and have a measured density of .5: They will not be stacked. We now have 83.33 boxes, of which 50 are light and 33.33 are heavy. If the 50 light boxes are stacked, regardless of whether the table is Fragile or Sturdy, it will resist 50 times. If the table is Fragile, it will resist 40% of the heavy boxes (13.33 times). Thus given that the table is Fragile and that 83.33 boxes are stacked, the table will resist 76% of the boxes (63.33 times). Similarly, given that the table is Sturdy and that 83.33 boxes are stacked, the table will resist 84% of the boxes (70 times), which is derived from resisting 60% of the heavy boxes (20 times) and all the light boxes (50 times). The likelihood ratio, given that a box has measured density of (.2 .3 .4), it is stacked on the

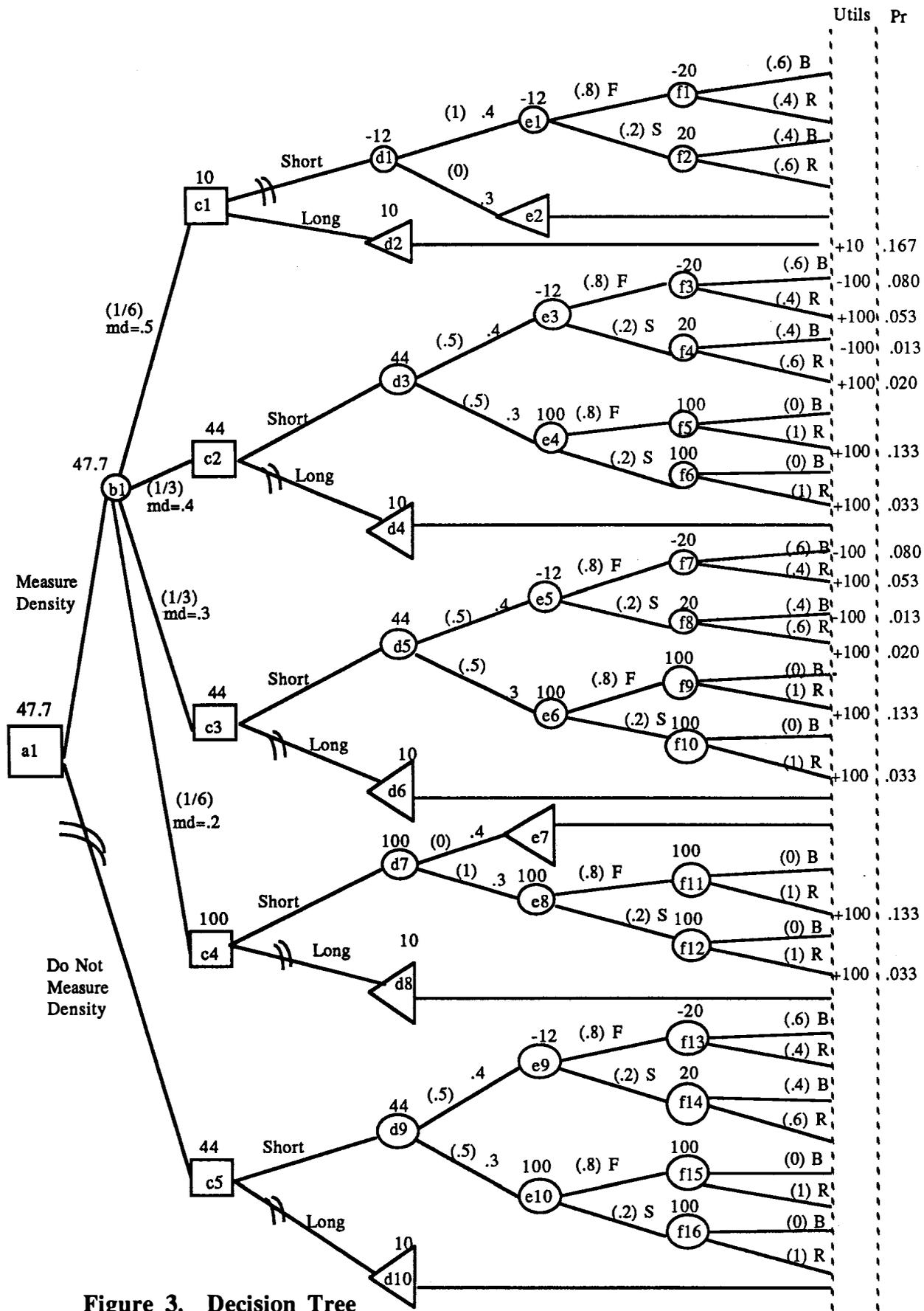

**Figure 3. Decision Tree**

Probabilities are in parentheses.  Pr: probability of taking entire path. (Pr=0 omitted)
md: measured density  F: Fragile  S: Sturdy  B: Breaks  R: Resists  True Density: .3 or .4

table, and the table resists, is:

$$L(Resists) = \frac{P(Resists/Fragile)}{P(Resists/Sturdy)}$$

$$= \frac{.76}{.84} = .905.$$

The likelihood ratio that the table breaks is:

$$L(Breaks) = \frac{P(Breaks/Fragile)}{P(Breaks/Sturdy)}$$

$$= \frac{.24}{.16} = 1.5.$$

In the transition from state 1 to state 2, it was observed that the table resisted. Multiply the prior odds of 4 by the appropriate likelihood ratio to get the posterior odds of 3.62. We can recover the probabilities by using the formula

$$P(Fragile/Observation) = \frac{O_{post}}{1+O_{post}} = .78.$$

The belief in state 2 that the table is Fragile has dropped from .80 to .78. If the table had been observed to break, the posterior odds would have gone to 6 and the belief that the table was Fragile would have gone to .857.

The decisions about density measurement and paths depend on the beliefs about the table's strength. If the table is in truth Sturdy, then at some time in the future we would expect that the robot would have inductively learned enough to change the decisions being made. At the present, the robot decides to measure every box and then chooses the short path for all boxes except when the density is measured as .5. If the table were known to be Sturdy, then the best decision would be always to choose the short path. Moreover, there would be no further gain from measurement, so the decision would be not to measure. What can be said about the point at which the robot's decisions will be changed as a result of inductive learning?

The decision tree shows that when measured density is .5, the decision to take the long or short path depends on the ex ected utility of outcomes at the chance node e1. The expected utility at e1 is -12. The decision will be changed to taking the short path if the expected utility at d1 is greater than +10. This can be represented as

$P(Fragile) EU(Fragile)$
$+ P(Sturdy) EU(Sturdy) > 10$

where $EU(Fragile)=-20$ is the expected utility at node f1 of a decision to stack a box, given that the box's density is .4 and that the table is Fragile. Similarly, at f2 we have $EU(Sturdy)=+20$. After solving for the two probabilities, we find that the inequality will be satisfied when

$P(Fragile) = .25$

$P(Sturdy) = .75.$

The decisions will be changed as soon as it is believed with probability .75 that the table is Sturdy. How long will that take?

If the table is Sturdy, then for the boxes with measured densities of {.2 .3 .4}, it has been shown that the table will resist 84% of the time while it will break 16% of the time. The average effect of applying values for the two likelihood ratios, $L(Resists)$ and $L(Breaks)$, in those proportions can be found by taking their geometric mean using .84 and .16 as weights. If the table is Sturdy, the prior odds will be multiplied by the average likelihood ratio

$$L(Ave)_{Sturdy} = (.905^{.84})(1.5^{.16}) = .9812.$$

If the table were actually Fragile, the weights would be .76 and .24 giving an average likelihood ratio

$$L(Ave)_{Fragile} = (.905^{.76})(1.5^{.24}) = 1.0217.$$

From the average likelihood ratio for a Sturdy table and the initial odds of 4, we can calculate that it will take 131 state transformations on the average before the odds go to 1/3, which corresponds to the belief of .75 that the table is sturdy:

$$4(.9812)^x = 1/3$$

$$x = 130.9$$

At that stage the robot will always choose the short path and will decide not to measure density.

The updating of probabilities can be done by any of several proposed procedures (Kim and Pearl 1983; Shacter 1986; Spiegelhalter 1986). In the present example, the most straightforward way would be to update directly the probabilities in N5 after observing the outcome at N7 and, similarly, to update the probabilities at N1 after observing the outcome at N12. The remaining probabilities can be directly calculated using the relations from the influence diagram to determine their values in the succeeding states.

## 5. Summary and Future Research

The T-BIL method integrates an essentially deductive learning method based on EBG with inductive learning methods from Bayesian decision theory. It takes as inputs a decision problem, a training example, and domain theory and constructs a generalization that explains why the example is a solution to the decision problem. The generalized explanation is represented as an influence diagram and is subject to the constraint that it be operational. It is used not only to explain events, but also as the basis for making sequential decisions. Once the decisions are made in the initial state, a transition to a new state occurs. In the new state the outcomes of each decision are observed and the explanation of observed events is revised using Bayesian updating. This procedure is iterative, thus providing adaptive stochastic decision making. Learning takes place in a stochastic environment.

The emphasis in T-BIL has been on the integration of symbolic and quantitative methods. Although there has been much discussion of the interdisciplinary nature of work in AI, there has in fact been very little exploitation of statistical and decision-analytic methods by AI workers. Perhaps surprisingly, there also has not been much migration of AI techniques into non-AI domains. As a colleague recently remarked, there are many conferences being given with the title "AI and X", where X is any other field. But when you look closely at the work being done and the people attending, the conference is mostly just X.

T-BIL uses tools that have been developed both within AI and in other domains. The key contribution from AI centers on the use of goal regression and an EBG-like approach to provide the context and to generate the explanation of the relationship between theory and fact. Bayesian decision theory, dynamic programing, and stochastic adaptive control theory have been the sources for the quantitative methods related to decision making, belief revision, and influence diagrams.

T-BIL has used these tools to provide four advantages over EBG:

- Explanations include probabilistic and statistical reasoning.

- Explanations are related to decisions that have utility to decision makers.

- Explanations and decisions are revised incrementally on the basis of observation reports.

- Disjuncts can explicitly be represented in the explanation.

The next steps in my research program will be to implement T-BIL as a working program and to find suitable real-world applications. Since T-BIL is a general technique that can deal with sequential decisions in a stochastic environment, it should be applicable to a wide variety of problems. For example, much laboratory work of a scientific nature involves examining data in order to decide which of several competing hypotheses is most strongly supported. Statistical packages require the user to interpret the statistical results and make the decision. A program built using T-BIL should be able to automate a substantial portion of that decision-making process.

In the world of finance, a real-time trading program should be able to analyze stock-price data and change its view of the world based on a continuous inflow of new data (Star 1986). Existing expert systems are not able to adapt quickly enough to changes in the environment to be useful in the financial arena. They also have difficulty dealing with statistical hypotheses. T-BIL should help us to solve some of these problems, thus moving us closer to the day when a financial program can act as an on-line real-time intelligent assistant that will analyze data, accept or reject various different statistical hypotheses, and recommend the most profitable decisions on the basis of its analysis.

Another problem area where T-BIL might prove to be useful involves the automatic diagnosis of computer network problems (Maxion 1986; Lin and Siewiorek 1986). Training examples based on specific types of faults could be used to generate more generalized explanations of the system's behavior. If data from error logs were automatically monitored, as certain patterns of error reports appear, the automatic diagnostic system could propose the most likely hypotheses and alert the operator as to a potential problem. As more data arrive, T-BIL would lead to modifications of the initial hypotheses. The particular actions recommended would depend in part on the cost associated with attempting the specific repairs given the uncertain knowledge about the true state of the world.

Both symbolic and quantitative computational techniques have proven themselves to provide very powerful problem-solving paradigms. An optimistic view of the interdisciplinary work being attempted suggests that the result of integrating symbolic and quantitative methods will be synergistic. It is certainly an interesting problem to be working on and one that is full of potential.